\documentclass[sigconf]{acmart}
\usepackage{multirow}
\AtBeginDocument{%
  \providecommand\BibTeX{{%
    \normalfont B\kern-0.5em{\scshape i\kern-0.25em b}\kern-0.8em\TeX}}}

\acmConference[AI4TV2019]{AI4TV 2019: 1\textsuperscript{st} International Workshop on AI for Smart TV Content Production, Access and Delivery}{October 21--25, 2019}{Nice, France}
\acmBooktitle{AI4TV 2019: 1\textsuperscript{st} International Workshop on AI for Smart TV Content Production, Access and Delivery, October 21--25, 2019, Nice, France}
\begin{document}

%%
%% The "title" command has an optional parameter,
%% allowing the author to define a "short title" to be used in page headers.
\title{Gender Representation in French Broadcast Corpora and Its Impact on ASR Performance}

%%
%% The "author" command and its associated commands are used to define
%% the authors and their affiliations.
%% Of note is the shared affiliation of the first two authors, and the
%% "authornote" and "authornotemark" commands
%% used to denote shared contribution to the research.
\author{Mahault Garnerin}
\affiliation{%
  \institution{Univ. Grenoble Alpes}
  \city{Grenoble}
  \country{France}}
\email{mahault.garnerin@univ-grenoble-alpes.fr}

\author{Solange Rossato}
\affiliation{%
  \institution{Univ. Grenoble Alpes}
  \city{Grenoble}
  \country{France}}
\email{solange.rossato@univ-grenoble-alpes.fr}

\author{Laurent Besacier}
\affiliation{%
  \institution{Univ. Grenoble Alpes}
  \city{Grenoble}
  \country{France}}
\email{laurent.besacier@univ-grenoble-alpes.fr}

%%
%% By default, the full list of authors will be used in the page
%% headers. Often, this list is too long, and will overlap
%% other information printed in the page headers. This command allows
%% the author to define a more concise list
%% of authors' names for this purpose.
\renewcommand{\shortauthors}{Garnerin and Rossato, et al.}

%%
%% The abstract is a short summary of the work to be presented in the
%% article.
\begin{abstract}
    This paper analyzes the gender representation in four major corpora of French broadcast. These corpora being widely used within the speech processing community, they are a primary material for training automatic speech recognition (ASR) systems. As gender bias has been highlighted in numerous natural language processing (NLP) applications, we study the impact of the gender imbalance in TV and radio broadcast on the performance of an ASR system. This analysis shows that women are under-represented in our data in terms of speakers and speech turns. We introduce the notion of speaker role to refine our analysis and find that women are even fewer within the Anchor category corresponding to prominent speakers. The disparity of available data for both gender causes performance to decrease on women. However this global trend can be counterbalanced for speaker who are used to speak in the media when sufficient amount of data is available.
    %canonical pronunciation.
\end{abstract}

%%
%% The code below is generated by the tool at http://dl.acm.org/ccs.cfm.
%% Please copy and paste the code instead of the example below.
%%


%%
%% Keywords. The author(s) should pick words that accurately describe
%% the work being presented. Separate the keywords with commas.

%% A "teaser" image appears between the author and affiliation
%% information and the body of the document, and typically spans the
%% page.
%\begin{teaserfigure}
%  \includegraphics[width=\textwidth]{sampleteaser}
%  \caption{Seattle Mariners at Spring Training, 2010.}
%  \Description{Enjoying the baseball game from the third-base
%  seats. Ichiro Suzuki preparing to bat.}
%  \label{fig:teaser}
%\end{teaserfigure}

%%
%% This command processes the author and affiliation and title
%% information and builds the first part of the formatted document.
\maketitle

\section{Introduction}
In recent years, gender has become a hot topic within the political, societal and research spheres. Numerous studies have been conducted in order to evaluate the presence of women in media, often revealing their under-representation, such as the Global Media Monitoring Project \cite{GMMP2015}. In the French context, the CSA\footnote{Conseil Sup{\'e}rieur de l'Audiovisuel: French equivalent of the American Federal Communications Commission or of the British Office of Communication} \cite{CSA2017} produces a report on gender representation in media on a yearly basis. The 2017 report shows that women represent 40\% of French media speakers, with a significant drop during high-audience hours (6:00-8:00pm) reaching a value of only 29\%. Another large scale study confirmed this trend with an automatic analysis of gender in French audiovisuals streams, highlighting a huge variation across type of shows \cite{doukhan2018open}.

Besides the social impact of gender representation, broadcast recordings are also a valuable source of data for the speech processing community. Indeed, automatic speech recognition (ASR) systems require large amount of annotated speech data to be efficiently trained, which leaves us facing the emerging concern about the fact that \textit{"AI artifacts tend to reflect the goals, knowledge and experience of their creators"} \cite{parsheera2018gendered}. Since we know that women are under-represented in media and that the AI discipline has retained a male-oriented focus \cite{wang2011aihalloffame}, we can legitimately wonder about the impact of using such data as a training set for ASR technologies. This concern is strengthened by the recent works uncovering gender bias in several natural language processing (NLP) tools such as \cite{bolukbasi2016man,caliskan2017semantics,vanmassenhove2018getting,tatman17-dialect-gender-effect}.

In this paper, we first highlight the importance of TV and radio broadcast as a source of data for ASR, and the potential impact it can have. We then perform a statistical analysis of gender representation in a data set composed of four state-of-the-art corpora of French broadcast, widely used within the speech community. Finally we question the impact of such a representation on the systems developed on this data, 
through the perspective of an ASR system.

\section{From gender representation in data to gender bias in AI}
\subsection{On the importance of data}
The ever growing use of machine learning in science has been enabled by several progresses among which the exponential growth of data available. The quality of a system now depends mostly on the quality and quantity of the data it has been trained on. If it does not discard the importance of an appropriate architecture, it reaffirms the fact that rich and large corpora are a valuable resource. Corpora are research contributions which do not only allow to save and observe certain phenomena or validate a hypothesis or model, but are also a mandatory part of the technology development. This trend is notably observable within the NLP field, where industrial technologies, such as Apple, Amazon or Google vocal assistants now reach high performance level partly due to the amount of data possessed by these companies \cite{chiu2018state}.

Surprisingly, as data is said to be ``the new oil", few data sets are available for ASR systems. The best known are corpora like TIMIT \cite{garofolo1993darpa}, Switchboard \cite{godfrey1992switchboard} or Fisher \cite{cieri2004fisher} which date back to the early 1990s. The scarceness of available corpora is justified by the fact that gathering and annotating audio data is costly both in terms of money and time. Telephone conversations and broadcast recordings have been the primary source of spontaneous speech used. 
%Challenge lately addressed : separation of sources, reverberation, noisy environment (CHiME & ASpIRE) 
%The latest broadcast corpus available on LDC\footnote{Linguistic Data Consortium : https://www.ldc.upenn.edu/language-resources} is the 2002 Rich Transcription Broadcast News & Conversational Telephone Speech released in 2004 and containing broadcast excerpts recorded in 1998 as well as excerpts from the Switchboard I and II corpora. The scarceness of available corpora is justified by the fact that gathering and annotating audio data is costly both in terms of money and time. Efforts are made to try to provide the community with more up-to-date data, such as the TEDLIUM corpus \cite{hernandez2018ted} or the Common Voice project\footnote{https://voice.mozilla.org/}.} 
Out of all the 130 audio resources proposed by LDC\footnote{Linguistic Data Consortium: \url{https://www.ldc.upenn.edu/language-resources}} to train automatic speech recognition systems in English, approximately 14\% of them are based on broadcast news and conversation. 
%Other resource are mainly focused on telephone speech (like Switchboard, CALLHOME, Fisher) or specific domain like ATIS (Air Travel Info Service). 
For French speech technologies, four corpora containing radio and TV broadcast are the most widely used: ESTER1 \cite{galliano2005ester}, ESTER2 \cite{galliano2009ester}, ETAPE \cite{gravier2012etape} and REPERE \cite{giraudel2012repere}. These four corpora have been built alongside evaluation campaigns and are still, to our knowledge, the largest French ones of their type available to date.    

\subsection{From data to bias}
The gender issue has returned to the forefront of the media scene in recent years and with the emergence of AI technologies in our daily lives, gender bias has become a scientific topic that researchers are just beginning to address. Several studies revealed the existence of gender bias in AI technologies such as face recognition (\textit{GenderShades} \cite{buolamwini2018gender}), NLP (word embeddings \cite{bolukbasi2016man} and semantics \cite{caliskan2017semantics}) and machine translation (\cite{prates2018genderbias_ml,vanmassenhove2018getting}). The impact of the training data used within these deep-learning algorithms is therefore questioned.

Bias can be found at different levels as pointed out by \cite{sun2019mitigatinggender}. \cite{crawford2017nips} defines bias as a skew that produces a type of harm. She distinguishes two types of harms that are allocation harm and representation harm. The allocation harm occurs when a system is performing better or worse for a certain group while representational harm contributes to the perpetuation of stereotypes. Both types of harm are the results of bias in machine learning that often comes from the data systems are trained on. Disparities in representation in our social structures is captured and reflected by the training data, through statistical patterns. The \textit{GenderShades} study is a striking example of what data disparity and lack of representation can produce: the authors tested several gender recognition modules used by facial recognition tools and found difference in error-rate as high as 34 percentage points between recognition of white male and black female faces. The scarce presence of women and colored people in training set resulted in bias in performance towards these two categories, with a strong intersectional bias. As written by~\cite{boyd2012critical} \textit{"A data set may have many millions of pieces of data, but this does not mean it is random or representative. To make statistical claims about a data set, we need to know where data is coming from; it is similarly important to know and account for the weaknesses in that data."} (p.668).

% \textcolor{cyan}{"Just because Big Data presents us with large quantities of data does not mean that methodological issues are no longer relevant. Understanding sample, for example, is more important now than ever." \cite{boyd2012critical}}

Regarding ASR technology, little work has explored the presence of gender bias within the systems and no consensus has been reached. \cite{adda2005speech} found that speech recognizers perform better on female voice on a broadcast news and telephone corpus. They proposed several explanations to this observation, such as the larger presence of non-professional male speech in the broadcast data, implying a less prepared speech for these speakers or a more normative language and standard pronunciation for women linked to the traditional role of women in language acquisition and education. The same trend was observed by \cite{goldwater2010words}. More recently, \cite{tatman2017gender} discovered a gender bias within YouTube's automatic captioning system but this bias was not observed in a second study evaluating Bing Speech system and YouTube Automatic Captions on a larger data set \cite{tatman17-dialect-gender-effect}. However race and dialect bias were found. General American speakers and white speakers had the lowest error rate for both systems. If the better performance on General American speakers could be explained by the fact that they are all voice professionals, producing clear and articulated speech, but no explanation is provided for biases towards non-white speakers. 

Gender bias in ASR technology is still an open research question as no clear answer has been reached so far. It seems that many parameters are to take into account to achieve a general agreement. As we established the importance of TV and radio broadcast as a source of data for ASR, and the potential impact it can have, the following content of this paper is structured as this: we first describe statistically the gender representation of a data set composed of four state-of-the-art corpora of French broadcast, widely used within the speech community, introducing the notion of speaker's role to refine our analysis in terms of voice professionalism. We then question the impact of such a representation on a ASR system trained on these data. \cite{elloumi2018asr-pred}

\section{Methodology}

This section is organized as follows: we first present the data we are working on. In a second time we explain how we proceed to describe the gender representation in our corpus and introduce the notion of speaker's role. The third subsection introduces the ASR system and metrics used to evaluate gender bias in performance.

\subsection{Data presentation}

Our data consists of two sets used to train and evaluate our automatic speech recognition system. Four major evaluation campaigns have enabled the creation of wide corpora of French broadcast speech: ESTER1 \cite{galliano2005ester}, ESTER2 \cite{galliano2009ester}, ETAPE \cite{gravier2012etape} and REPERE \cite{giraudel2012repere}. These four collections contain radio and/or TV broadcasts aired between 1998 and 2013 which are used by most academic researchers in ASR. Show duration varies between 10min and an hour. As years went by and speech processing research was progressing, the difficulty of the tasks augmented and the content of these evaluation corpora changed. %alongside this aim
ESTER1 and ESTER2 mainly contain prepared speech such as broadcast news, whereas ETAPE and REPERE consists also of debates and entertainment shows, spontaneous speech introducing more difficulty in its recognition. 

Our training set contains 27,085 speech utterances produced by 2,506 speakers, accounting for approximately 100 hours of speech. Our evaluation set contains 74,064 speech utterances produced by 1,268 speakers for a total of 70 hours of speech.
Training data by show, medium and speech type is summarized in Table~\ref{tab:train_show_list} and evaluation data in Table~\ref{tab:test_show_list}. Evaluation data has a higher variety of shows with both prepared (P) and spontaneous (S) speech type (accented speech from African radio broadcast is also included in the evaluation set).

\begin{table}
\caption{Training data description}
 \label{tab:train_show_list}
\centering
\begin{tabular}{lrcc}
\toprule
Show               & Duration  & Medium & Type \\
\midrule
BFM Story          & 25h 36min & TV     & P \\
France Info Infos  & 11h 23min & Radio  & P \\
France Inter Infos & 42h 45min & Radio  & P \\
LCP Infos          & 10h 6min  & TV     & P \\
RFI Infos          & 1h 49min  & Radio  & P \\
Top Questions      & 7h 59min  & TV     & P \\
\midrule
\textbf{Total}     & \textbf{99h 38min} & -   & P \\
\bottomrule
\end{tabular}
\end{table}
\begin{table}
\caption{Evaluation data description}
 \label{tab:test_show_list}
\centering
\begin{tabular}{lrcc}
\toprule
Show                    & Duration  & Medium  & Type \\
\midrule
Africa1                 & 1h 21min  & Radio   & P \\
Comme On Nous parle     & 2h 14min  & Radio   & S \\
Culture et Vous         & 1h 16min  & TV      & S \\
La Place du Village     & 1h 24min  & TV      & S \\
Le Masque et la Plume   & 4h 12min  & Radio   & S \\
Pile et Face            & 7h 52min  & TV      & P \\
Planete Showbiz         & 1h 12min  & TV      & S \\
RFI Infos               & 24h 14min & Radio   & P \\
RTM Infos               & 22h 0min  & Radio   & P \\
Service Public          & 2h 30min  & Radio   & S \\
TVME Infos              &  57min    & Radio   & P \\
Un Temps de Pauchon     & 1h 31min  & Radio   & S \\
\midrule
\textbf{Total}          & \textbf{70h43min}  & -  & - \\
\bottomrule
\end{tabular}
\end{table}

\subsection{Methodology for descriptive analysis of gender representation in training data}
\label{sec:method-data}

We first describe the gender representation in training data. Gender representation is measured in terms of number of speakers, number of utterances (or speech turns), and turn lengths (descriptive statistics are given in Section~\ref{sec:rslt_gend_rep}). Each speech turn was mapped to its speaker in order to associate it with a gender.

As pointed out by the CSA report \cite{CSA2017}, women presence tends to be marginal within the high-audience hours, showing that women are represented but less than men and within certain given conditions. It is clear that a small number of speakers is responsible for a large number of speech turns. Most of these speakers are journalists, politicians, presenters and such, who are representative of a show. Therefore, we introduce the notion of speaker's role to refine our exploration of gender disparity, following studies which quantified women's presence in terms of role. Within our work, we define the notion of speaker role by two criteria specifying the speaker's on-air presence, namely the number of speech turns and the cumulative duration of his or her speaking time in a show.
Based on the available speech transcriptions and meta-data, we compute for each speaker the number of speech turns uttered as well as their total length. We then use the following criteria to define speaker's role: a speaker is considered as speaking often (respectively seldom) if he/she accumulates a total of turns higher (respectively lower) than 1\% of the total number of speech turns in a given show. The same process is applied to identify speakers talking for a long period from those who do not. We end up with two salient roles called Anchors and Punctual speakers:
\begin{itemize}
    \item the \textbf{Anchor speakers (A)} are above the threshold of 1\% for both criteria, meaning they are intervening often and for a long time thus holding an important place in interaction; 
    \item the \textbf{Punctual speakers (PS)} on the contrary are below the threshold of 1\% for both the total number of turns and the total speech time.
\end{itemize}

These roles are defined at the show level. They could be roughly assimilated to the categorization ``host/guest'' in radio and TV shows. Anchors could be described as professional speakers, producing mostly prepared speech, whereas Punctual speakers are more likely to be ``everyday people". The concept of speaker's role  makes sense at both sociological and technical levels. An Anchor speaker is more likely to be known from the audience (society), but he or she will also likely have a professional (clear) way of speaking (as mentioned by \cite{adda2005speech} and \cite{tatman17-dialect-gender-effect}), as well as a high number of utterances, augmenting the amount of data available for a given gender category.

\subsection{Gender bias evaluation procedure of an ASR system performance}

\subsubsection{ASR system}

The ASR system used in this work is described in \cite{elloumi2018asr-pred}. It uses the KALDI toolkit \cite{povey2011kaldi}, following a standard Kaldi recipe. The acoustic model is based on a hybrid HMM-DNN architecture and trained on the data summarized in Table~\ref{tab:train_show_list}. Acoustic training data correspond to 100h of non-spontaneous speech type (mostly broadcast news) coming from both radio and TV shows.
%aired between 1998 and 2013. 
A 5-gram language model is trained from several French corpora (3,323M words in total\footnote{from EUbookshop, TED2013, Wit3, GlobalVoices, Gigaword, Europarl-v7, MultiUN, OpenSubtitles2016, DGT, News Commentary, News WMT, LeMonde, Trames, Wikipedia and transcriptions of the training dataset.}) using SRILM toolkit \cite{stolcke2002srilm}. The pronunciation model is developed using the lexical resource BDLEX~\cite{de1998bdlex} as well as automatic grapheme-to-phoneme (G2P)\footnote{Available at: \url{http://lia.univ-avignon.fr/chercheurs/bechet/download\_fred.html}} transcription to find pronunciation variants of our vocabulary (limited to 80K). 
It is important to re-specify here, for further analysis, that our Kaldi pipeline follows speaker adaptive training (SAT) where we train and decode using speaker adapted features (fMLLR-adapted features) in per-speaker mode.  It is well known that speaker adaptation acts as an effective procedure to reduce mismatch between training and evaluation conditions \cite{leggetter1995maximum,povey2011kaldi}. 

\subsubsection{Evaluation}
\label{sec:method-wer}

Word Error Rate (WER) is a common metric to evaluate ASR performance. It is measured as the sum of errors (insertions, deletions and substitutions) divided by the total number of words in the reference transcription. 
As we are investigating the impact on performance of speaker's gender and role, we computed the WER for each speaker at the episode (show occurrence) level. Analyzing at such granularity allows us to avoid large WER variation that could be observed at utterance level (especially for short speech turns) but also makes possible to get several WER values for a given speaker, one for each occurrence of a show in which he/she appears on. Speaker's gender\footnote{Non-gendered (NA) speakers correspond mainly to overlapping speech or to some really short speech turns attributed to unknown speaker for which gender was not provided.} was provided by the meta-data and role was obtained using the criteria from Section~\ref{sec:method-data} computed for each show. This enables us to analyze our results across gender and role categories which was done using Wilcoxon rank sum tests also called Mann-Whitney U test (with $\alpha$= 0.001) \cite{mann1947test}. The choice of a Wilcoxon rank sum test and not the commonly used t-test is motivated by the non-normality of our data.

\section{Results}

\subsection{Descriptive analysis of gender representation in training data}

\subsubsection{Gender representation}
\label{sec:rslt_gend_rep}
\begin{table}[th]
 \caption{Gender representation in training data}
 \label{tab:spk_speechtime}
 \centering
 \begin{tabular}{lccc}
	\toprule
	& Female    & Male          & NA        \\
	\midrule
	\multirow{2}{*}{\textbf{Speakers}} 
	 & 831      & 1637          & 38        \\
     & (33.16\%)   & (65.32\%)       & (1.52\%)    \\
    \midrule
    \multirow{2}{*}{\textbf{Speech time}}
     & 22h 30min  & 75h 30min    & 1h 40min  \\
     & (22.57\%)   & (75.75\%)   & (1.68\%)        \\
	\bottomrule
  \end{tabular}
\end{table}

As expected, we observe a disparity in terms of gender representation in our data (see Table~\ref{tab:spk_speechtime}). Women represent 33.16\% of the speakers, confirming the figures given by the GMMP report \cite{GMMP2015}. However, it is worth noticing that women account for only 22.57\% of the total speech time, which leads us to conclude that women also speak less than men.

\subsubsection{Speaker's role representation}

Table~\ref{tab:roles_speechtime} presents roles' representation in training data and shows that despite the small number of Anchor speakers in our data (3.79\%), they nevertheless concentrate 35.71 \% of the total speech time.

\begin{table}[th]
 \caption{Roles representation in training data}
 \label{tab:roles_speechtime}
 \centering
 \begin{tabular}{lccc}
	\toprule
	            & Anchors (A)   & Punctual (PS)    & Others (O)  \\
	\midrule
	\multirow{2}{*}{\textbf{Speakers}} 
	            & 95           & 2325                      & 86        \\
                & (3.79\%)     & (92.78\%)                 & (3.43\%)   \\
    \hline
    \multirow{2}{*}{\textbf{Speech time}}
                & 35h 36min     & 49h 25min                 & 14h 39min  \\
                & (35.71\%)     & (49.59\%)                 & (14.70\%)   \\
	\bottomrule	
  \end{tabular}
\end{table}

\subsubsection{Role and gender interaction} 

\begin{table}
 \caption{Train data (percentages are calculated within role's categories)}
 \label{tab:train_data_v2}
\centering
\begin{tabular}{lcccc}
\toprule
          &  Role & F             & M             & NA            \\
\midrule
 \multirow{4}{*}{\textbf{Speakers}}
    & \multirow{2}{*}{\textbf{A}}
                & 28           & 63            & 4             \\
             &  & (29.47\%)    & (66.32\%)     & (4.21\%)      \\

    & \multirow{2}{*}{\textbf{PS}}
	            & 783          & 1510          & 32            \\
              & & (33.68\%)    & (64.95\%)     & (1.37\%)      \\
\midrule
 \multirow{4}{*}{\textbf{Speech time}}
    & \multirow{2}{*}{\textbf{A}}
                & 7h26min      & 26h 29min     & 1h 40min      \\
             &  & (20.89\%)    & (74.86\%)     & (4.25\%)       \\

    & \multirow{2}{*}{\textbf{PS}}
	            & 12h 21min    & 37h 00min     & 4min         \\
             &  & (25.00\%)    & (74.86\%)     & (0.14\%)      \\
\bottomrule
\end{tabular}
\end{table}

When crossing both parameters, we can observe that the gender distribution is not constant throughout roles. Women represent 29.47\% of the speakers within the Anchor category, even less than among the Punctual speakers. Their percentage of speech is also smaller. When calculating the average speech time uttered by a female Anchor, we obtain a value of 15.9 min against 25.2 min for a male Anchor, which suggests that even within the Anchor category men tend to speak more. This confirms the existence of gender disparities within French media. It corroborates with the analysis of the CSA \cite{CSA2017}, which shows that women were less present during high-audience hours. Our study shows that they are also less present in important roles. 
These results legitimate our initial questioning on the impact of gender balance on ASR performance trained on broadcast recordings.

\subsection{Performance (WER) analysis on evaluation data}

\subsubsection{Impact of gender on WER}
As explained in Section~\ref{sec:method-wer}, WER is the sum of errors divided by the number of words in the transcription reference. The higher the WER, the poorer the system performance. Our 70h evaluation data contains a large amount of spontaneous speech and is very challenging for the ASR system trained on prepared speech: we observe an overall average WER of 42.9\% for women and 34.3\% for men. This difference of WER between men and women is statistically significant (med(M)~=~25\%; med(F)~=~29\%; U~=~709040;~p-value~<~0.001).

\begin{figure}[t]
  \centering
  \includegraphics[width=\linewidth]{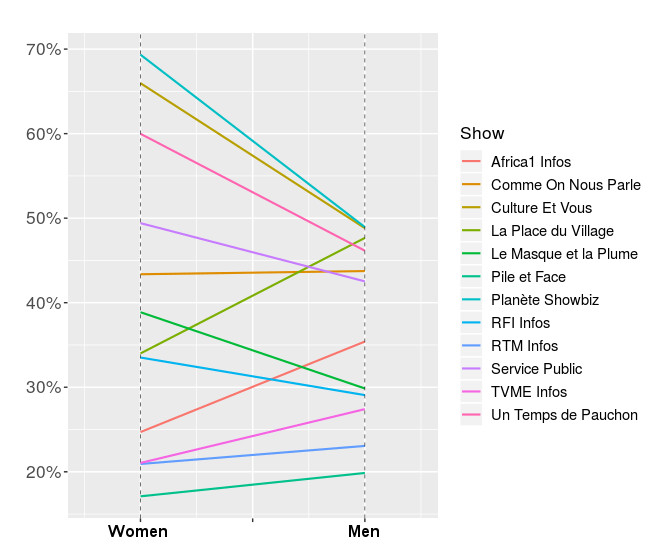}
  \caption{WER scores by show and gender - 70h evaluation set (WER being an error-rate, the smaller the better. If performance were always better for one gender, we would expect each line to have the same slope direction)}
  \label{fig:wer_gender_by_show}
\end{figure}

However, when observing gender differences across shows, no clear trend can be identified, as shown in Figure~\ref{fig:wer_gender_by_show}. For shows like \emph{Africa1 Infos} or \emph{La Place du Village}, we find an average WER lower for women than for men, while the trend is reversed for shows such as \emph{Un Temps de Pauchon} or \emph{Le Masque et la Plume}. The disparity of the results depending on the show leads us to believe that other factors may be entangled within the observed phenomenon.

\subsubsection{Impact of role on WER}

Speaker's role seems to have an impact on WER: we obtain an average WER of 30.8\% for the Anchor speakers and 42.23\% for the Punctual speakers. This difference is statistically significant with a p-value smaller than $10^{-14}$ (med(A)~=~21\%;~med(P)~=~31\%;~U~=~540,430;~p-value~<~0.001) .

\subsubsection{Role and gender interaction}
\begin{figure}[t]
  \centering
  \includegraphics[width=\linewidth]{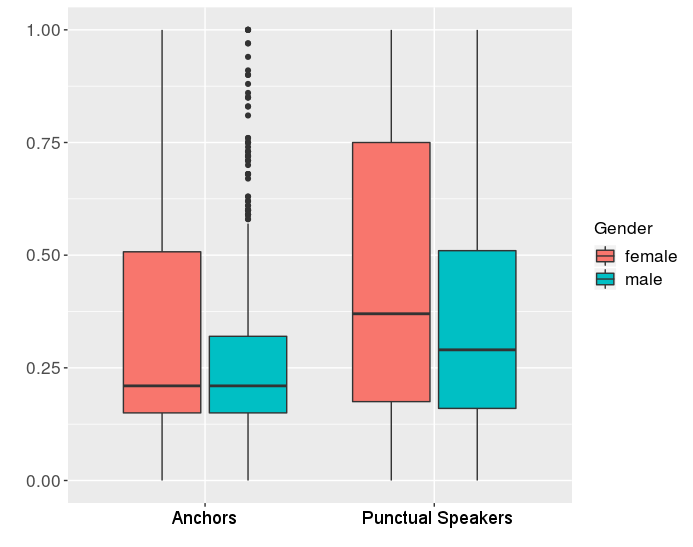}
  \caption{WER distribution by role and gender - 70h evaluation set}
  \label{fig:wer_gender_roles}
\end{figure}

Figure~\ref{fig:wer_gender_roles} presents the WER distribution (WER being obtained for each speaker in a show occurrence) according to the speaker's role and gender. It is worth noticing that the gender difference is only significant within the Punctual speakers group. The average WER is of 49.04\% for the women and 38.56\% for the men with a p-value smaller than $10^{-6}$ (med(F)~=~39\%; med(M)~=~29\%; U~=~251,450; p-value~<~0.001), whereas it is just a trend between male and female Anchors (med(F)~=~21\%; med(M)~=~21\%; U~=~116,230; p-value~=~0.173).
This could be explained by the quantity of data available per speaker. 

\subsubsection{Speech type as a third entangled factor?}

In order to try to explain the observed variation in our results depending on shows and gender (Figure~\ref{fig:wer_gender_by_show}), we add the notion of speech type to shed some light on our results. \cite{adda2005speech} and \cite{tatman2017gender} suggested that the speaker professionalism, associated with clear and hyper-articulated speech could be an explaining factor for better performance.  

\begin{figure}[t]
  \centering
  \includegraphics[width=\linewidth]{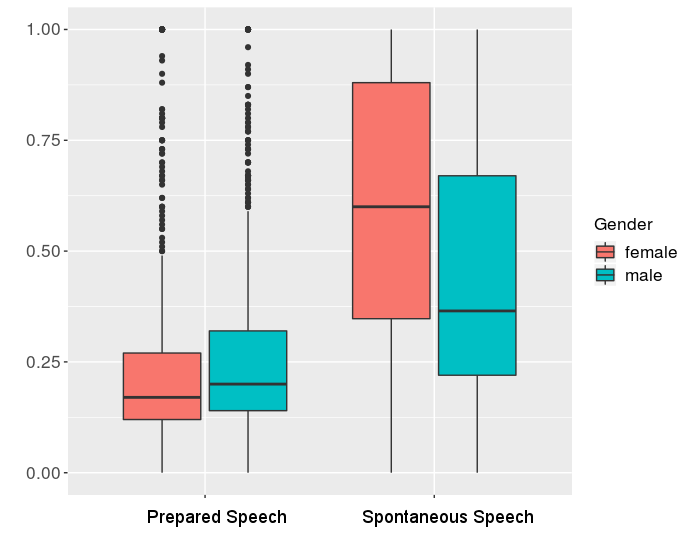}
  \caption{WER distribution by type of speech and gender - 70h evaluation set}
  \label{fig:wer_type_speech}
\end{figure}

Based on our categorization in prepared speech (mostly news reports) and spontaneous speech (mostly debates and entertainment shows), we cross this parameter in our performance analysis. As shown on Figure~\ref{fig:wer_type_speech}, these results confirm the inherent challenge of spontaneous speech compared to prepared speech. WER scores are similar between men and women when considering prepared speech (med(F)~=~18\%; med(M)~=~21\%; U~=~217,160; p-value~=~0.005) whereas they are worse for women (61.29\%) than for men (46.51\%) with p-value smaller than $10^{-14}$ for the spontaneous speech type (med(F)~=~61\%; med(M)~=~37\%; U~=~153,580; p-value~<~0.001). 

\section{Discussion}

We find a clear disparity in terms of women presence and speech quantity in French media. Our data being recorded between 1998 and 2013, we can expect this disparity to be smaller on more recent broadcast recordings, especially since the French government displays efforts toward parity in media representation. 
One can also argue that even if our analysis was conducted on a large amount of data it does not reach the exhaustiveness of large-scale studies such as the one of \cite{doukhan2018open}. Nonetheless it does not affect the relevance of our findings, because if real-world gender representation might be more balanced today, these corpora are still used as training data for AI systems.

The performance difference across gender we observed corroborates (on a larger quantity and variety of language data produced by more than 2400 speakers) the results obtained by \cite{tatman2017gender} on isolated words recognition. However the following study on read speech does not replicate these results. Yet a performance degradation is observed across dialect and race~\cite{tatman17-dialect-gender-effect}. \cite{adda2005speech} found lower WER for women than men on broadcast news and conversational telephone speech for both English and French. The authors suggest that gender stereotypes associated with women role in education and language acquisition induce a more normative elocution. We observed that the higher the degree of normativity of speech the smaller the gender difference. No significant gender bias is observed for prepared speech nor within the Anchor category. Even if we do not find similar results with lower WER for women than men, we obtained a median WER smaller for women on prepared speech and equal to the male median WER for the Anchor speakers. 

Another explanation could be the use of adaptation within the pipeline. Most broadcast programs transcription systems have a speaker adaptation step within their decoding pipeline, which is the case for our system. An Anchor speaker intervening more often would have a larger quantity of data to realize such adaptation of the acoustic model. On the contrary, Punctual speakers who appear scarcely in the data are not provided with the same amount of adaptation data. Hence we can hypothesize that gender performance difference observed for Punctual speakers is due to the fact that female speech is further from the (initial non-adapted) acoustic model as it was trained on unbalanced data (as shown in Table~\ref{tab:spk_speechtime}). Considering that Punctual speakers represent 92.78\% of the speakers, this explains why gender difference is significant over our entire data set. A way to confirm our hypothesis would be to reproduce our analysis on WER values obtained without using speaker adapted features at the decoding step.

When decoding prepared speech (hence similar to the training data), no significant difference is found in WER between men and women, revealing that the speaker adaptation step could be sufficient to reach same performance for both genders. But when decoding more spontaneous speech, there is a mismatch with the initial acoustic model (trained on prepared speech). Consequently, the speaker adaptation step might not be enough to recover good ASR performance, especially for women for whom less adaptation data is available (see Section 4.2.3).

\section{Conclusion}

This paper has investigated gender bias in ASR performance through the following research questions: i) what is the proportion of men and women in French radio and TV media data ? ii) what is the impact of the observed disparity on ASR performance ? iii) is this as simple as a problem of gender proportion in the training data or are other factors entangled ? Our contributions are the following:

\begin{itemize}
    \item Descriptive analysis of the broadcast data used to train our ASR system confirms the already known disparity, where 65\% of the speakers are men, speaking more than 75\% of the time. 
    \item When investigating WER scores according to gender, speaker's role and speech type, huge variations are observed. We conclude that gender is clearly a factor of variation in ASR performance, with a WER increase of 24\% for women compared to men, exhibiting a clear gender bias. 
    \item Gender bias varies across speaker's role and speech spontaneity level. Performance for Punctual speakers respectively spontaneous speech seems to reinforce this gender bias with a WER increase of 27.2\% respectively 31.8\% between male and female speakers.
\end{itemize}

We found that an ASR system trained on unbalanced data regarding gender produces gender bias performance. Therefore, in order to create fair systems it is necessary to take into account the representation problems in society that are going to be encapsulated in the data. Understanding how women under-representation in broadcast data can lead to bias in ASR performances is the key to prevent re-implementing and reinforcing discrimination already existing in our societies. This is in line with the concept of ``Fairness by Design" proposed by \cite{abbasi2018make}.

Gender, race, religion, nationality are all characteristics that we deem unfair to classify on, and these ethical standpoints needs to be taken into account in systems' design. Characteristics that are not considered as relevant in a given task can be encapsulated in data nonetheless, and lead to bias performance. Being aware of the demographic skews our data set might contain is a first step to track the life cycle of a training data set and a necessary step to control the tools we develop.

%% The next two lines define the bibliography style to be used, and
%% the bibliography file.
\bibliographystyle{ACM-Reference-Format}
\bibliography{my_bib}

\end{document}